\documentclass[final,1p,times]{elsarticle}

\usepackage{amssymb}

\usepackage{algorithm2e}
\usepackage{amsfonts}
\usepackage{bm}
\usepackage{physics}
\usepackage{url}
\usepackage{xcolor}

\SetAlCapSty{}
\SetKwComment{Comment}{/* }{ */}

\begin{document}

\begin{frontmatter}



\title{4Ward: a Relayering Strategy for Efficient Training of Arbitrarily Complex Directed Acyclic Graphs}


\author[inst1]{Tommaso Boccato}
\ead{tommaso.boccato@uniroma2.it}
\author[inst1]{Matteo Ferrante}
\author[inst1]{Andrea Duggento}
\author[inst1,inst2]{Nicola Toschi}

\affiliation[inst1]{organization={Department of Biomedicine and Prevention, University of Rome Tor Vergata},
            city={Rome},
            country={Italy}}
\affiliation[inst2]{organization={A.A. Martinos Center for Biomedical Imaging and Harvard Medical School},
            city={Boston},
            country={USA}}

\begin{abstract}
Thanks to their ease of implementation, multilayer perceptrons (MLPs) have become ubiquitous in deep learning applications. The graph underlying an MLP is indeed multipartite, i.e. each layer of neurons only connects to neurons belonging to the adjacent layer. In contrast, in vivo brain connectomes at the level of individual synapses suggest that biological neuronal networks are characterized by scale-free degree distributions or exponentially truncated power law strength distributions, hinting at potentially novel avenues for the exploitation of evolution-derived neuronal networks. In this paper, we present ``4Ward'', a method and Python library capable of generating flexible and efficient neural networks (NNs) from arbitrarily complex directed acyclic graphs. 4Ward is inspired by layering algorithms drawn from the graph drawing discipline to implement efficient forward passes, and provides significant time gains in computational experiments with various Erdős-Rényi graphs. 4Ward not only overcomes the sequential nature of the \textit{learning matrix} method, by parallelizing the computation of activations, but also addresses the scalability issues encountered in the current state-of-the-art and provides the designer with freedom to customize weight initialization and activation functions. Our algorithm can be of aid for any investigator seeking to exploit complex topologies in a NN design framework at the microscale.
\end{abstract}




\end{frontmatter}


\section{Introduction}\label{sec:intro}

\begin{figure}
    \centering
    \includegraphics[width=\textwidth]{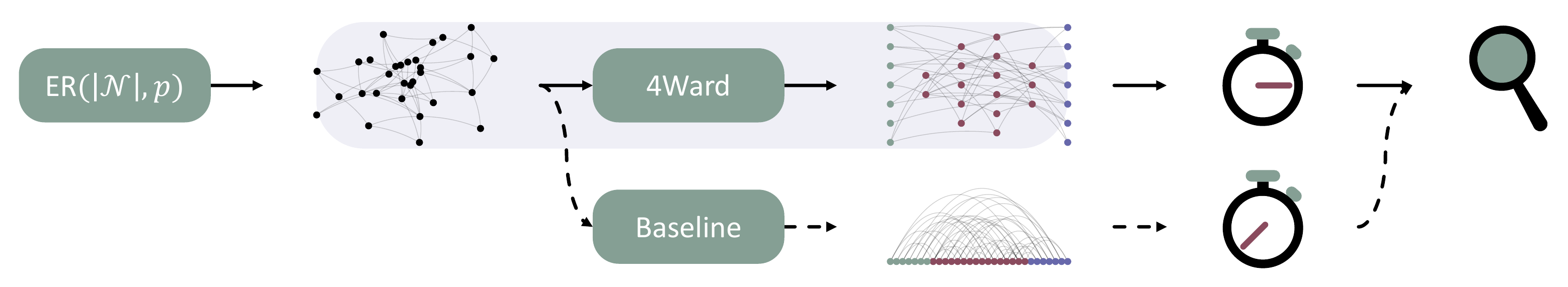}
    \caption{Time complexity evaluation of the 4Ward algorithm. The process starts with the generation of Erdős-Rényi (ER) \cite{erdHos1960evolution} graphs from a predetermined parameter grid. Each undirected network is then converted into a DAG following a random heuristic. As highlighted in purple, 4Ward is able to process the DAGs produced and return deployable NNs. Finally, the execution times of the forward passes of the networks are measured and compared with those obtained from the sequential baseline \cite{monteiro_model_2016}.}
    \label{fig:overview}
\end{figure}

Thanks to their ease of implementation, multilayer perceptrons (MLPs) have become ubiquitous in deep learning (DL) applications. Virtually all state-of-the-art neural architectures (e.g., convolutional neural networks \cite{2014arXiv1409.1556S,7780459}, variational autoencoders \cite{DBLP:journals/corr/KingmaW13}, transformers \cite{NIPS2017_3f5ee243}) include MLPs among their functional blocks. Amongst other advantages, a forward pass in a MLP reduces to a series of matrix multiplications, guaranteeing computational efficiency.

The graph underlying an MLP, however, is multipartite. In other words, each layer of neurons only connects to neurons belonging to the adjacent layer. In this context, studies conducted on in vivo brain connectomes, at the level of individual synapses, suggest that biological neuronal networks are characterized by scale-free degree distributions \cite{varshney_structural_2011} or exponentially truncated power law strength distributions \cite{scheffer_connectome_2020}. These architectural differences between biological and artificial neural networks (NNs) suggest potentially novel avenues for the exploitation of evolution‐derived neuronal networks as artificial NNs by relaxing the topological constraints of MLPs.

In the current DL landscape, there is no satisfactory tool that allows to efficiently experiment, on the microscale, with analog (i.e., nonspiking) NNs based on arbitrarily complex topologies. Current strategies to convert complex networks into NNs can be divided into three main categories: topology-dependent, layer-specific and generalist.

Topology-dependent approaches include strategies in which NNs are generated by randomly rewiring a fraction of the edges of a MLP \cite{simard_fastest_2005,erkaymaz_performance_2012,erkaymaz_impact_nodate,erkaymaz_impact_2016,erkaymaz_performance_2017}. Typically, these models can be trained with standard backpropagation; however, the range of usable topologies is constrained by the random interpolation procedure used on the original MLP. Additionally, the strategy presented in \cite{basili_evolving_2007} is compatible with Barabási–Albert graphs \cite{RevModPhys.74.47} only, and the difficulties related to the extension of backpropagation toward nonlayered graphs is overcome through the use of evolutionary algorithms for training the networks.

Layer-specific approaches include \cite{mocanu_scalable_2018}, where the focus is placed on the interlayer connections of MLPs. The procedure employed (sparse evolutionary training - SET), is capable of inducing sparsity in the MLPs’ bicliques by pruning the weakest connections. The produced topologies, however, are a result of the SET procedure and, hence, not defined \textit{a priori}. Also the framework presented in \cite{you_graph_2020} allows the definition of a trainable topology through a tool called relational graph. Still, the expressiveness of relational graphs is limited to the bipartite graphs that underlie an MLP.

Finally, the generalist category is mostly represented by the \textit{learning matrix method} \cite{monteiro_model_2016,platt_computational_2019}. The learning matrix is an augmented adjacency matrix on which forward and backward operations are performed, i.e. it includes additional rows and columns that contain the state information required to compute the forward/backward passes. The framework accepts directed acyclic graphs (DAGs) represented as upper triangular matrices; however, neuron activations are computed sequentially w.r.t. the network nodes. As a consequence, the method does not scale with the network size. The recent work by Stier et al., \texttt{deepstruct} \cite{STIER2019107,STIER2022100193}, faces a similar issue. In the model's forward pass, neural activations are decomposed using a series of matrix multiplications, which are defined based on a pre-calculated partitioning of the NN nodes. This decomposition leads to a reduced number of matrix multiplications when the graph has a density close to zero, demonstrating sublinear dependence on the number of neurons; otherwise, the required number of multiplications can exhibit a quadratic dependence.

Outside the three discussed categories, several NNs architectures have computational graphs defined on a macroscopic scale \cite{9010992,NEURIPS2019_d010396c,roberts_deep_2019}. In these works, nodes typically represent differentiable functional blocks (e.g., linear layers, convolutions), while the edges are tensors. Neural architecture search, in both its ``classical'' \cite{DBLP:conf/iclr/ZophL17} and differentiable \cite{DBLP:conf/iclr/LiuSY19,gu_dots_2021} variants, also falls within this body of research. Although these works have achieved remarkable results in computer vision and natural language processing, they do not address the requirements for which microscopic-scale models were designed: transforming and combining features without a specific spatial relationship.

In order to overcome the mentioned microscale-related difficulties, in this paper, we present ``4Ward'', a method and Python library\footnote{\url{https://github.com/BoCtrl-C/forward}} capable of generating flexible and efficient NNs from arbitrary DAGs. The method minimizes the number of matrix multiplications required to perform a forward pass, and is inspired by layering algorithms drawn from the \textit{graph drawing} discipline to implement efficient forward passes. The 4Ward functional blocks are deployable as modules compatible with PyTorch \cite{NEURIPS2019_9015}. An overview of the experimental protocol developed to evaluate the time complexity of the 4Ward algorithm is reported in Figure \ref{fig:overview}.

\section{Methods}\label{sec:methods}

The computational graph is a powerful formalism used to represent NNs. Nodes are typically associated with function-variable pairs, $(f_v, a_v),\ v \in \mathcal{N} \subset \mathbb{N}$, while each edge is associated with a weight $w_{uv},\ (u, v) \in \mathcal{E} \subset \mathcal{N} \times \mathcal{N}$. When the values of the variables associated with the predecessors of node $v$ are fixed, its activation can be computed through:
\begin{equation}
    a_v = f_v\Bigg(\sum_{u : (u, v) \in \mathcal{E}} w_{uv}a_u\Bigg)
\end{equation}
Graph sources (i.e., $\{v : \{(u, v) \in \mathcal{E}\} = \emptyset\}$) are called input nodes since their role is to provide input values for subsequent function evaluations. Similarly, sinks (i.e., $\{v : \{(v, u) \in \mathcal{E}\} = \emptyset\}$) are called output nodes since their activations correspond to the result of previous function evaluations.

Directed graphs with no directed cycles are the backbone of feedforward NNs. DAGs are of particular interest as they admit at least one topological ordering, that is a linear ordering such that $\forall\ (u, v) \in \mathcal{E}\ u < v$; this implies the possibility of computing the output values of the graph, given a set of values for the input nodes, evaluating functions in the same order expressed by the topological one. However, this straightforward implementation of the forward pass (i.e., the process that leads to the evaluation of the output nodes) is inefficient due to its sequential nature. All nodes whose predecessors' values are already available can, in principle, be evaluated in parallel.

MLPs, whose underlying graph is a chain of bicliques, are particularly well suited for an efficient implementation of the forward pass. All activations of a layer, $\bm{a}^l$, are computed simultaneously from the activations of the previous layer, $\bm{a}^{l - 1}$, through matrix multiplication\footnote{The overall time complexity of the forward pass depends on the matrix multiplication algorithm.}:
\begin{equation}\label{eq:mlp-layer-f}
    \bm{a}^l = \sigma(W^l\bm{a}^{l - 1})
\end{equation}
where $\bm{a}^l \in \mathbb{R}^\abs{L_l}$, $L_l$ is the set of nodes in layer $l$, $W^l \in \mathbb{R}^{\abs{L_l} \times \abs{L_{l - 1}}}$ represents the weights and $\sigma$ denotes the activation function. While the computational graph of an MLP is characterized by a specific topology for which a node partition is immediately identifiable, it is worth noting that the same does not hold for arbitrary DAGs. The latter require relayering algorithms to implement efficient forward passes.

\subsection{4Ward}\label{sec:4ward}

Let the layering $\mathcal{L} = \{L_0, \dots, L_{H - 1}\}$ be a partition of $\mathcal{N}$ such that $\forall\ (u, v) \in \mathcal{E}\ u \in L_i, v \in L_j \Longrightarrow i < j$. As in the MLP case, all functions associated with the $l$-th layer can be evaluated in parallel since the required input values are already available at computation time; this is guaranteed by the definition of layering. Consequently, equation\eqref{eq:mlp-layer-f} can be generalized as follows:
\begin{equation}\label{eq:4ward-layer-f}
    \bm{a}^l = \sigma(W^l\bm{x}^l)
\end{equation}
where $\bm{x}^l = [\dots, a_v, \dots]^T,\ v \in P_l$ and $P_l = \{u : \forall\ v \in L_l\ (u, v) \in \mathcal{E}\}$. In other words, $\bm{x}^l$ contains the activations of the predecessors (i.e., $P_l$) of nodes belonging to layer $l$. Importantly, in $W^l \in \mathbb{R}^{\abs{L_l} \times \abs{P_l}}$, all entries corresponding to $(\mathcal{N} \times \mathcal{N}) \setminus \mathcal{E}$ must be 0 to preserve the given DAG topology; this constraint is enforced in the implementation phase through the \textit{mask trick}. A major difference from equation \eqref{eq:mlp-layer-f} lies in the ability of nodes belonging to layer $l$ to draw information from the subnetwork $\cup_{i = 0}^{l - 1} L_i$.

Under the assumption that the overall time complexity of a forward pass is dominated by $H$, optimizing the graph partitioning becomes crucial\footnote{It is also important to take the impact of memory accesses and data transfers between devices into account.}. In this context, the \textit{longest-path layering algorithm} \cite{10.5555/3122413,10.1007/3-540-45848-4_2} can be employed. A layering algorithm can be seen as a mapping between graphs and layerings: $\mathcal{G} \mapsto \mathcal{L}$; with $\mathcal{G}=(\mathcal{N}, \mathcal{E})$. In general, DAGs admit more than one layering; the longest-path algorithm, however, returns one of those of minimum height (i.e., $H$). While it is possible for the minimum height layering to be non-unique, in accordance with the above assumption, different layerings of same height lead to the same time complexity. It is worth noting that topological orderings can be seen as a particular case of layering in which $H = \mathcal{O}(\abs{\mathcal{N}})$. Instead, in minimum height layerings, $H$ is topology-dependent.

\subsection{Implementation}\label{sec:implementation}

\begin{figure}
    \centering
    \includegraphics[width=\textwidth]{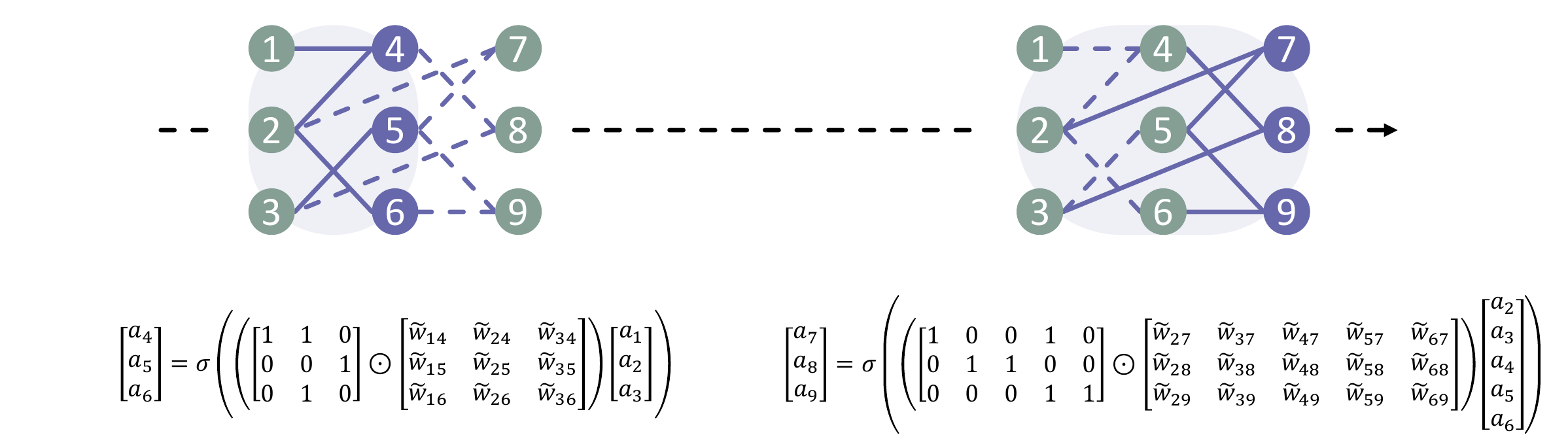}
    \caption{Forward pass computation for an example NN built through 4Ward (from left to right). \textbf{Left}: computation of the activations of layer 1. \textbf{Right}: computation of the activations of layer 2. The subnetworks involved are highlighted through the purple overlay. Computations follow $\bm{a}^l = \sigma((M^l \odot \tilde{W}^l)\bm{x}^l)$, in accordance with \eqref{eq:4ward-layer-f}.}
    \label{fig:fw-pass-example}
\end{figure}

We chose PyTorch \cite{NEURIPS2019_9015} as our reference machine learning framework for its ease of use in research prototyping, its automatic differentiation engine, Autograd, and its popularity among scientists. Implementing the forward pass in a PyTorch \texttt{Module} is immediate, with the exception of preventing the gradient from updating the zero entries of all $W^l$. PyTorch tensors, indeed, can be marked as learnable or nonlearnable; however, the topology of the DAG which is object of relayering needs to be preserved.

To circumnavigate this issue, we propose a strategy which we term the mask trick. It consists of masking the fully-learnable tensor $\tilde{W}^l$; this is accomplished by pointwise multiplication $W^l = M^l \odot \tilde{W}^l$. Specifically, the $l$-th mask matrix is defined as:
\begin{equation}
    m_{uv} =
    \begin{cases}
    1\quad \text{if}\ (u, v) \in \mathcal{E}\\
    0\quad \text{otherwise}
    \end{cases}
\end{equation}
where $m_{uv}$ denotes the entry of $M^l$ linked to edge $(u, v)$. The partial derivative of a loss function $l$ w.r.t. $\tilde{w}_{uv}$ immediately follows:
\begin{align}
    \pdv{l(h_{\bm{w}}(\bm{a}^0), y)}{\tilde{w}_{uv}} &= \pdv{l}{h_{\bm{w}}}\pdv{h_{\bm{w}}}{\tilde{w}_{uv}}\nonumber\\
    &= \pdv{l}{h_{\bm{w}}}\pdv{h_{\bm{w}}}{w_{uv}}\pdv{w_{uv}}{\tilde{w}_{uv}}\nonumber\\
    &= \pdv{l}{h_{\bm{w}}}\pdv{h_{\bm{w}}}{w_{uv}}m_{uv}
\end{align}
where $h_{\bm{w}}$ denotes the NN parameterized by $\{w_{uv},\ (u, v) \in \mathcal{E}\}$, $(\bm{a}^0, y)$ is the generic labeled training sample and $w_{uv} = m_{uv}\tilde{w}_{uv}$. As a consequence, $m_{uv} = 0\ \Longrightarrow\ \pdv{l}{\tilde{w}_{uv}} = 0$.

The pseudocode in Algorithms \ref{alg:4ward-init} and \ref{alg:4ward-fw} provides a detailed description of the 4Ward implementation. Figure \ref{fig:fw-pass-example}, instead, shows an example forward pass computation. In Algorithm \ref{alg:4ward-init}, $longest\_path()$ wraps the layering algorithm, $push\_sources()$ moves the source nodes within $L_0$\footnote{The considered layering algorithm, by default, places each node in the highest possible layer.} and $init()$ initializes the weights. The $init()$ function accepts $M^l$ as an argument to allow the use of sophisticated initialization methods (e.g., Kaiming Uniform \cite{7410480}) that take into account the in-degree of nodes. Biases can be easily included in the framework by adding a column to $\tilde{W}^l$, e.g. replacing $M^l$ with $[M^l, \bm{1}]$ and $\bm{x}^l$ with $[(\bm{x}^l)^T, 1]^T$. Extending $\bm{x}^l$ to matrix $X^l$ allows 4Ward to work with batches.\\

\begin{algorithm}[H]
\caption{4Ward instantiation.}\label{alg:4ward-init}
$\mathcal{L} \gets longest\_path(\mathcal{G})$\;
$\mathcal{L} \gets push\_sources(\mathcal{G}, \mathcal{L})$\;
\For{$l \gets 1$ to $H - 1$}{
    $P^l \gets \emptyset$\;
    \For{$v \in L_l$}{
        $P^l \cup \{u : (u, v) \in \mathcal{E}\}$
    }
    $M^l \gets \bm{0}$\;
    $\tilde{W}^l \gets \bm{0}$\;
    \For{$v \in L_l$}{
        \For{$u \in \{u : (u, v) \in \mathcal{E}\}$}{
            $m_{uv} \gets 1$
        }
    }
    $\tilde{W}^l \gets init(\tilde{W}^l, M^l)$
}
\end{algorithm}

\begin{algorithm}[H]
\caption{4Ward forward pass.}\label{alg:4ward-fw}
\For{$l \gets 1$ to $H - 1$}{
    $\bm{a}^l \gets \sigma((M^l \odot \tilde{W}^l)\bm{x}^l)$\Comment*[r]{the mask trick}
}
\end{algorithm}

\subsection{Baseline}

In addition to 4Ward, we developed a baseline for the experimental protocol described in Section \ref{sec:experiments}. The baseline is a PyTorch-compatible extension of the learning matrix method \cite{monteiro_model_2016,platt_computational_2019} which implements the sequential forward pass discussed above. The baseline pseudocode is shown in Algorithms \ref{alg:baseline-init} and \ref{alg:baseline-fw}. Algorithm \ref{alg:baseline-init} reports the model instantiation while Algorithm \ref{alg:baseline-fw} describes how the forward pass is computed. In both algorithms, $\bm{w}_v$ denotes the weight vector of node $v$, and $\bm{a}^v$ stores the activations of the $v$'s predecessors. Both vectors share the same dimension: $\abs{\{u : (u, v) \in \mathcal{E}\}}$.\\

\begin{algorithm}[H]
\caption{Baseline instantiation.}\label{alg:baseline-init}
sources $\gets \{v : \{(u, v) \in \mathcal{E}\} = \emptyset\}$\;
\For{$v \in \mathcal{N} \setminus \textup{sources}$}{
    $\bm{w}_v \gets \bm{0}$\;
    $init(\bm{w}_v)$\;
}
\end{algorithm}

\begin{algorithm}[H]
\caption{Baseline forward pass.}\label{alg:baseline-fw}
\For{$v \in \mathcal{N} \setminus \textup{sources}$}{
    $a_v \gets \sigma(\bm{w}_v^T\bm{a}^v)$
}
\end{algorithm}

\subsection{Deepstruct}\label{sec:deepstruct}

We also incorporated the \texttt{deepstruct} methodology, mentioned in Section \ref{sec:intro}, into our experiments. In the high sparsity regime, the efficiency of this methodology, in terms of time complexity, depends on how the computation of activations is decomposed within the forward pass. Similar to 4Ward, \texttt{deepstruct} initializes the input DAG by partitioning it into a layering of minimum height. However, layers are exploited in a different manner during the forward pass. To compute the activations of a specific layer, $\bm{a}^l$, \texttt{deepstruct} follows the equation:
\begin{equation}\label{eq:ds}
    \bm{a}^l = \sigma\bigg(\sum_{j = 0}^{l - 1} W^{j \mapsto l}\bm{a^j}\bigg)
\end{equation}
According to \eqref{eq:ds}, each layer preceding layer $l$ contributes additively to the computation of $\bm{a}^l$ with its own activations (i.e., $\bm{a}^j$) through the matrix multiplication $W^{j \mapsto l}\bm{a^j}$, where $W^{j \mapsto l}$ represents the synaptic weights connecting neurons in layer $j$ to neurons in layer $l$. In contrast to \eqref{eq:4ward-layer-f}, where activations are computed all at once, the summation in \eqref{eq:ds} involves $l$ matrix multiplications. It is important to note that, in the worst case, $l$ corresponds to the number of nodes preceding $L_l$ (i.e., $\abs{\cup_{j = 0}^{l - 1}L_{j}}$, where each $L_j$ contains only one node).

We provide a detailed description of the \texttt{deepstruct} implementation of \eqref{eq:ds} in Algorithm \ref{alg:ds-fw}. The first for loop iterates over the layers, while the second loop performs the summation in \eqref{eq:ds}. Partial activations are cumulatively stored in the $\bm{a}^l$ variable. As in Algorithm \ref{alg:4ward-fw}, $\tilde{W}^{j \mapsto l}$, $M^{j \mapsto l}$, and $\sigma$ represent the raw weights between layers $j$ and $l$, their mask, and the activation function, respectively. If a mask is completely empty, the associated multiplication operation is skipped. When $H \sim \abs{\mathcal{N}}$, the time complexity of the algorithm, in terms of matrix multiplications, becomes quadratic in the number of nodes.

\begin{algorithm}[H]
\caption{\texttt{deepstruct} forward pass.}\label{alg:ds-fw}
\For{$l \gets 1$ to $H - 1$}{
    $\bm{a}^l \gets \bm{0}$\;
    \For{$j \gets 0$ to $l - 1$}{
        \If{$M^{j \mapsto l} \neq \bm{0}$}{
            $\bm{a}^l \gets \bm{a}^l + (M^{j \mapsto l} \odot \tilde{W}^{j \mapsto l})\bm{a}^j$\;
        }
    }
    $\bm{a}^l \gets \sigma(\bm{a}^l)$\;
}
\end{algorithm}

\subsection{Experiments}\label{sec:experiments}

Evaluating the time complexity of Algorithm \ref{alg:4ward-fw} analytically, w.r.t. a traditional set of graph attributes (e.g., $\abs{\mathcal{N}}$ and $\abs{\mathcal{E}}$), is not viable, especially in view of the dependency on low-level operations (e.g., matrix multiplications) and, most importantly, on the topology-dependency of $H$ and $\tilde{W}^l$. Instead, we rely on an empirical evaluation of the execution time of our proposed forward pass. Specifically, in a set of experiments, we monitored how long it took neural networks created in accordance with the same generative model to process a predetermined number of batches. Notably, the actual content of a batch is irrelevant for the purposes of time evaluation. Hence, all input tensors were filled with 1s.

We set the Erdős-Rényi (ER) model \cite{erdHos1960evolution} as our reference generator. Given a graph of size $\abs{\mathcal{N}}$, the ER algorithm randomly places the network edges by sampling from $\abs{\mathcal{E}} = \frac{\abs{\mathcal{N}}(\abs{\mathcal{N}} - 1)}{2}$ i.i.d. Bernoulli distributions, $\mathcal{B}(1, p)$. Subsequently, the largest connected component of each network was extracted and converted into a DAG following the procedure in \cite{9010992}.

For completeness, we repeated the exact same temporal assessment for all the $(\abs{\mathcal{N}}, p)$ pairs belonging to a 2D grid defined over the ER's size-probability parameter space, averaging, for each parameter pair, several execution times linked to a fixed number of ER generation seeds: $\mathbb{E}\big[\Delta T_{\scriptscriptstyle\abs{\mathcal{N}}, p}^{\scriptscriptstyle\bullet}\big]$ where $\bullet$ can take on the values $4W$, $SQ$ or $DS$ depending on the methodology considered - 4Ward, sequential baseline or \texttt{deepstruct}. The mean gains were calculated by running the same experimental protocol on $\mathbb{E}\big[\Delta T_{\scriptscriptstyle\abs{\mathcal{N}}, p}^{\scriptscriptstyle SQ}/\Delta T_{\scriptscriptstyle\abs{\mathcal{N}}, p}^{\scriptscriptstyle 4W}\big]$ and $\mathbb{E}\big[\Delta T_{\scriptscriptstyle\abs{\mathcal{N}}, p}^{\scriptscriptstyle DS}/\Delta T_{\scriptscriptstyle\abs{\mathcal{N}}, p}^{\scriptscriptstyle 4W}\big]$. All experiments where run on a workstation equipped with an Intel Xeon Gold 6326 CPU, 512 GB of RAM and an NVIDIA RTX A6000 GPU (driver version 510.47, CUDA version 11.6).

\section{Results}\label{sec:results}

We implemented the experimental protocol described in Section \ref{sec:experiments} while setting $\abs{\mathcal{N}} \in \{2^6, 2^7, 2^8, 2^9, 2^{10}\}$ and $p \in \{0.2, 0.4, 0.6, 0.8, 1.0\}$. For each $(\abs{\mathcal{N}}, p)$ pair, 30 graphs were generated starting from different random seeds. We tested three batch sizes: 32, 128 and 512; all computations involved a sequence of 100 batches. The experiments were repeated for all the implementations (i.e., baseline, \texttt{deepstruct} and 4Ward).

\begin{figure}
    \centering
    \includegraphics[width=\textwidth]{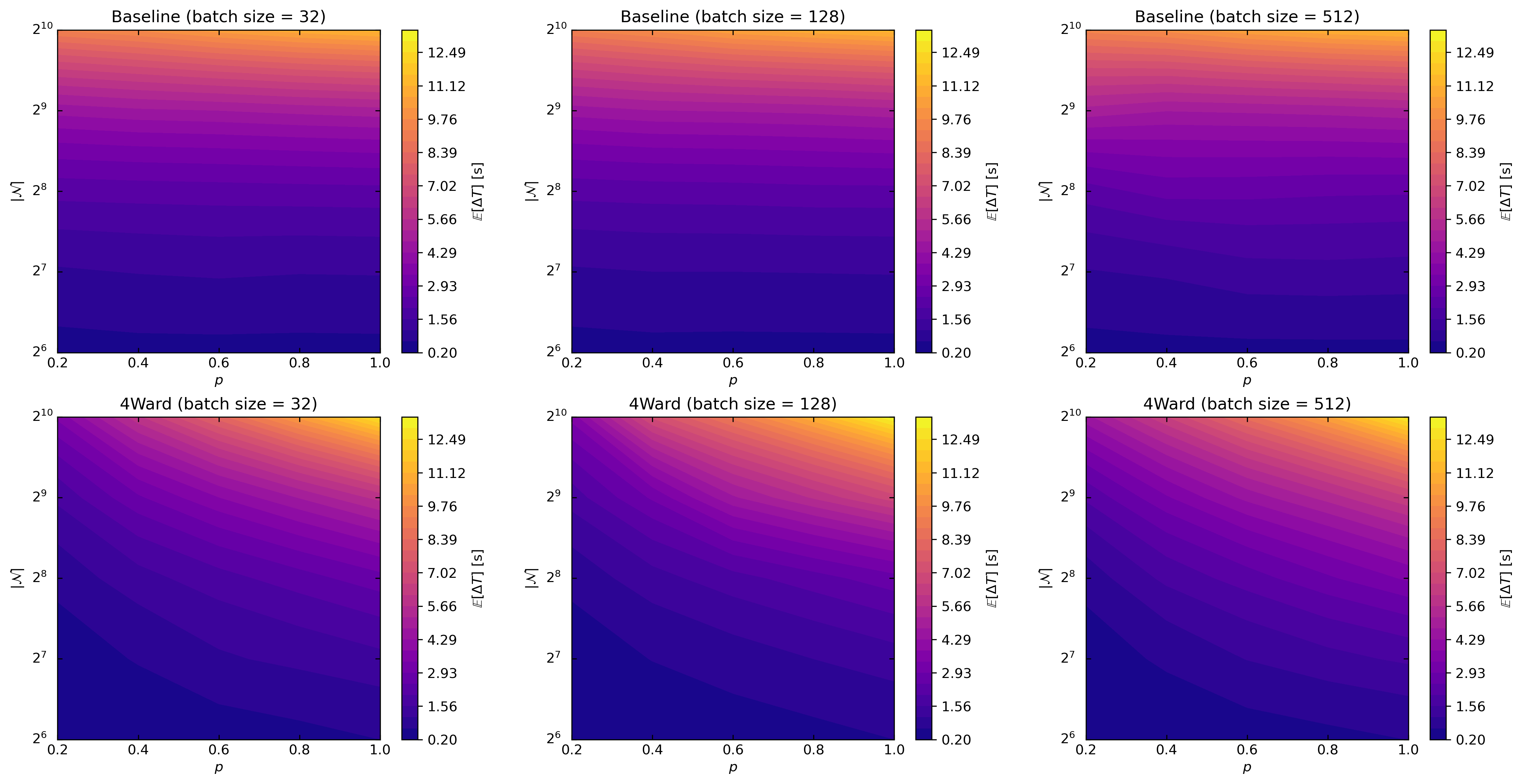}
    \caption{Mean execution times in seconds. Timings were examined on 100 forward passes. \textbf{Top}: baseline. \textbf{Bottom:} 4Ward. Each column refers to a specific batch size. Plots span the investigated ER parameter space. Network sizes are represented on the $y$-axis, while probabilities are represented on the $x$-axis. Elapsed times are color-coded.}
    \label{fig:times}
\end{figure}

The computed mean execution times (baseline and 4Ward) are reported in Figure \ref{fig:times}. Once the probability $p$ is fixed, function $\mathbb{E}\big[\Delta T_{\scriptscriptstyle\abs{\mathcal{N}}, p}\big]$ grows with the network size; this applies to both the baseline and our method. However, the two implementations behave differently as the ER probability varies. The execution time of the baseline's forward pass does not appear to depend on $p$, while that of 4Ward increases with $p$. In other words, the gradient of $\mathbb{E}\big[\Delta T_{\scriptscriptstyle\abs{\mathcal{N}}, p}^{\scriptscriptstyle SQ}\big]$ w.r.t. $p$ and $\abs{\mathcal{N}}$ is almost vertical, the one of $\mathbb{E}\big[\Delta T_{\scriptscriptstyle\abs{\mathcal{N}}, p}^{\scriptscriptstyle 4W}\big]$ points toward the top-right corner of the investigated parameter space. No differences emerged from the data collected for the different batch sizes.

\begin{figure}
    \centering
    \includegraphics[width=\textwidth]{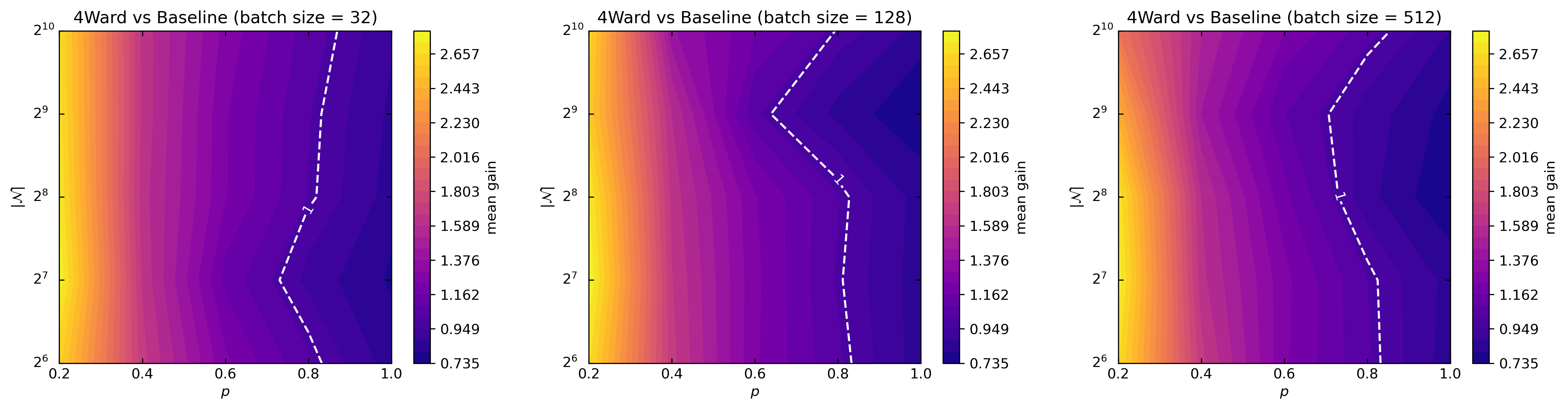}
    \includegraphics[width=\textwidth]{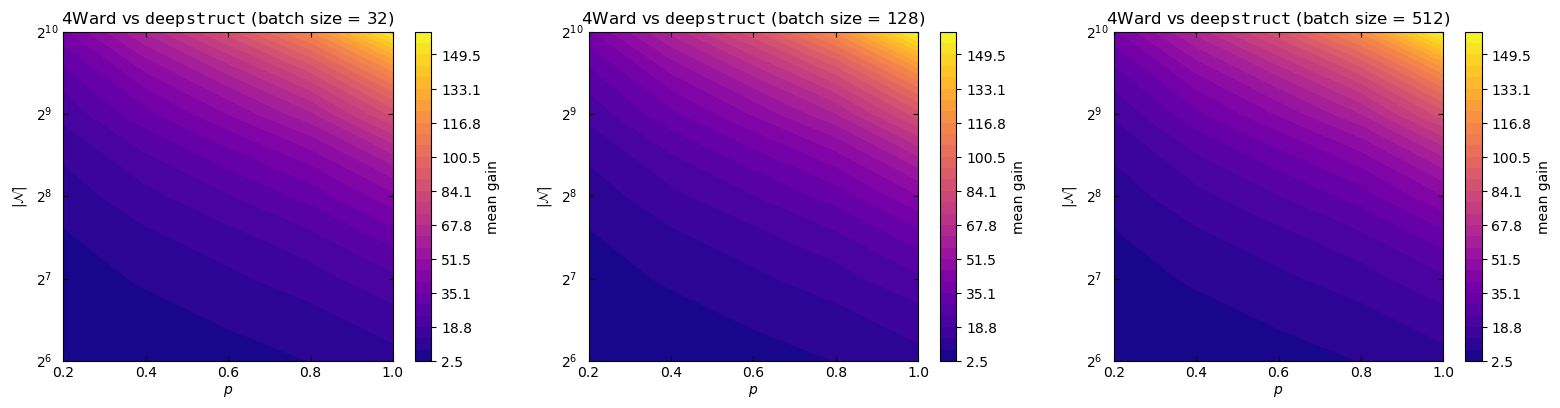}
    \caption{Mean gains. \textbf{Top}: 4Ward vs. baseline. \textbf{Bottom}: 4Ward vs. \texttt{deepstruct}. Each column refers to a specific batch size. Plots span the same size-probability parameter space of Figure \ref{fig:times}. Gain values are color-coded. The isolines corresponding to the unity gain are denoted in white.}
    \label{fig:gains}
\end{figure}

Figure \ref{fig:gains} shows the results obtained in terms of mean gain (Section \ref{sec:experiments}). In the top panel, the gradient of the metric points toward the left boundary of the parameter space. The isolines corresponding to the unity gain vertically cut the tested spaces in a low-density region (i.e., $p \sim 0.8$). 4Ward decreases computation time by up to a factor 2.80, with higher gains at higher sparsity levels. In the bottom panel, instead, gradients point toward the top-right corners of the drawn parameter spaces, and therefore present a non-zero horizontal component. In this comparison, mean gains span across a minimum of 2.47 and a maximum of 160.37. Minor differences emerge for the different batch sizes.

\section{Discussion}

Our results verified the assumption that the overall time complexity of a forward pass is dominated by $H$. For the baseline, the result can be directly inferred from the mean times reported in Section \ref{sec:results}, which depend only on the number of nodes that characterizes the graphs. Our sequential implementation is indeed a special case of 4Ward in which the layering $\mathcal{L}$ always has one node per hidden layer (i.e., $H \sim \abs{\mathcal{N}}$). The same conclusion can be reached for 4Ward (see \ref{sec:height-attenuation}). The left panel of Figure \ref{fig:t-attrs} reports function $\mathbb{E}\big[H_{\scriptscriptstyle\abs{\mathcal{N}}, p}\big]$, which demonstrates in how many layers, on average, a graph corresponding to a precise $(\abs{\mathcal{N}}, p)$ pair is partitioned by the longest-path layering algorithm. Unsurprisingly, $\mathbb{E}\big[\Delta T_{\scriptscriptstyle\abs{\mathcal{N}}, p}^{\scriptscriptstyle 4W}\big]$ and $\mathbb{E}\big[H_{\scriptscriptstyle\abs{\mathcal{N}}, p}\big]$ exhibit the same behavior. The batch size, which was set to values commonly used in DL literature, has little impact on the analyzed execution times, potentially due to the comparatively high computational power of the A6000 unit.

The most distinctive feature of 4Ward, however, is its efficiency, as shown by the gain analysis of Section \ref{sec:results}. According to Figure \ref{fig:gains}, the mean gain function computed on the times of 4Ward and the baseline, $\mathbb{E}\big[\Delta T_{\scriptscriptstyle\abs{\mathcal{N}}, p}^{\scriptscriptstyle SQ}/\Delta T_{\scriptscriptstyle\abs{\mathcal{N}}, p}^{\scriptscriptstyle 4W}\big]$, presents a series of vertical bands characterized by the same gain. The mean gain increases as the ER probability decreases. In other words, it is particularly advantageous to rely on the forward pass defined by our library when dealing with sparse neural networks. An undirected network is defined to be sparse when $\abs{\mathcal{E}} \ll \frac{\abs{\mathcal{N}}(\abs{\mathcal{N}} - 1)}{2}$, and $\mathbb{E}[\abs{\mathcal{E}}] = p\frac{\abs{\mathcal{N}}(\abs{\mathcal{N}} - 1)}{2}$ for the ER model. In our experiments, the highest percentage gain recorded is 280\% ($p = 0.2$), and it can reasonably be assumed that this figure  would increase further with increasing sparsity. In the examined density regime, the baseline and 4Ward become equivalent when the DAG is fully-connected, because in this case it only admits one topological ordering in which each node is connected to the entire subnetwork that precedes it. This is experimentally confirmed in Figure \ref{fig:gains} (aside from a minor shift to the left in unity gain isolines potentially due to low-level implementation differences). The comparison between 4Ward and \texttt{deepstruct} yielded even more impressive results, showing a remarkable highest percentage gain of approximately 16000\%. Moreover, within the tested parameter space, \texttt{deepstruct} consistently exhibited forward pass times that were at least 2.4 times slower than those of 4Ward. Notably, the plots demonstrate diagonal gain ``isobands'', indicating that mean gains are influenced by both the size of the network, $\abs{\mathcal{N}}$, and the ER probability, $p$. Specifically, as network size and density increase, the mean gains also increase. We hypothesize that the two models will perform similarly along a curve that lies entirely within the narrow unexplored region of the parameter space defined by $p < 0.2$. At the lowest density regime tested (i.e., $p = 0.2$), the multiplication operations performed by \texttt{deepstruct} already result in a significant slowdown compared to 4Ward. Lastly, it is important to acknowledge that slight differences in the implementation of the two architectures might have introduced a minor bias in the measured timings.

\section{Conclusions}

In this paper we have presented and released 4Ward, a scalable network generator capable of converting arbitrary DAGs into complex NNs. 4Ward effectively addresses the sequential limitations of the \textit{learning matrix} method and resolves scalability concerns  by introducing a novel parallelization method for computing activations. Furthermore, 4Ward provides the designer with freedom to customize weight initialization and activation functions. The modules produced within the library can be trained using all PyTorch optimizers (e.g., Adam \cite{DBLP:journals/corr/KingmaB14}, stochastic gradient descent with momentum \cite{pmlr-v28-sutskever13}) due to their compatibility with the machine learning framework. 4Ward provides significant computational time gains and can be of aid for any investigator seeking to exploit complex topologies in a NN design framework at the microscale.

\section*{Acknowledgements}

This work is supported and funded by: \#NEXTGENERATIONEU (NGEU); the Ministry of University and Research (MUR); the National Recovery and Resilience Plan (NRRP); project MNESYS (PE0000006, to NT) - \textit{A Multiscale integrated approach to the study of the nervous system in health and disease} (DN. 1553 11.10.2022); the MUR-PNRR M4C2I1.3 PE6 project PE00000019 Heal Italia (to NT); the NATIONAL CENTRE FOR HPC, BIG DATA AND QUANTUM COMPUTING, within the spoke \textit{``Multiscale Modeling and Engineering Applications''} (to NT); the European Innovation Council (Project CROSSBRAIN - Grant Agreement 101070908, Project BRAINSTORM - Grant Agreement 101099355); the Horizon 2020 research and innovation Programme (Project EXPERIENCE - Grant Agreement 101017727). Tommaso Boccato is a PhD student enrolled in the National PhD in Artificial Intelligence, XXXVII cycle, course on Health and Life Sciences, organized by Università Campus Bio-Medico di Roma.

\appendix

\section{Height Attenuation}\label{sec:height-attenuation}

We define two new graph attributes useful to understand the temporal performance of 4Ward. The graph \textit{height}, $H$, tells us in how many layers a layering of minimum height partitions the graph. The \textit{height attenuation}, instead, is defined as $\frac{\abs{\mathcal{N}}}{H}$.

\begin{figure}
    \centering
    \includegraphics[width=.66\textwidth]{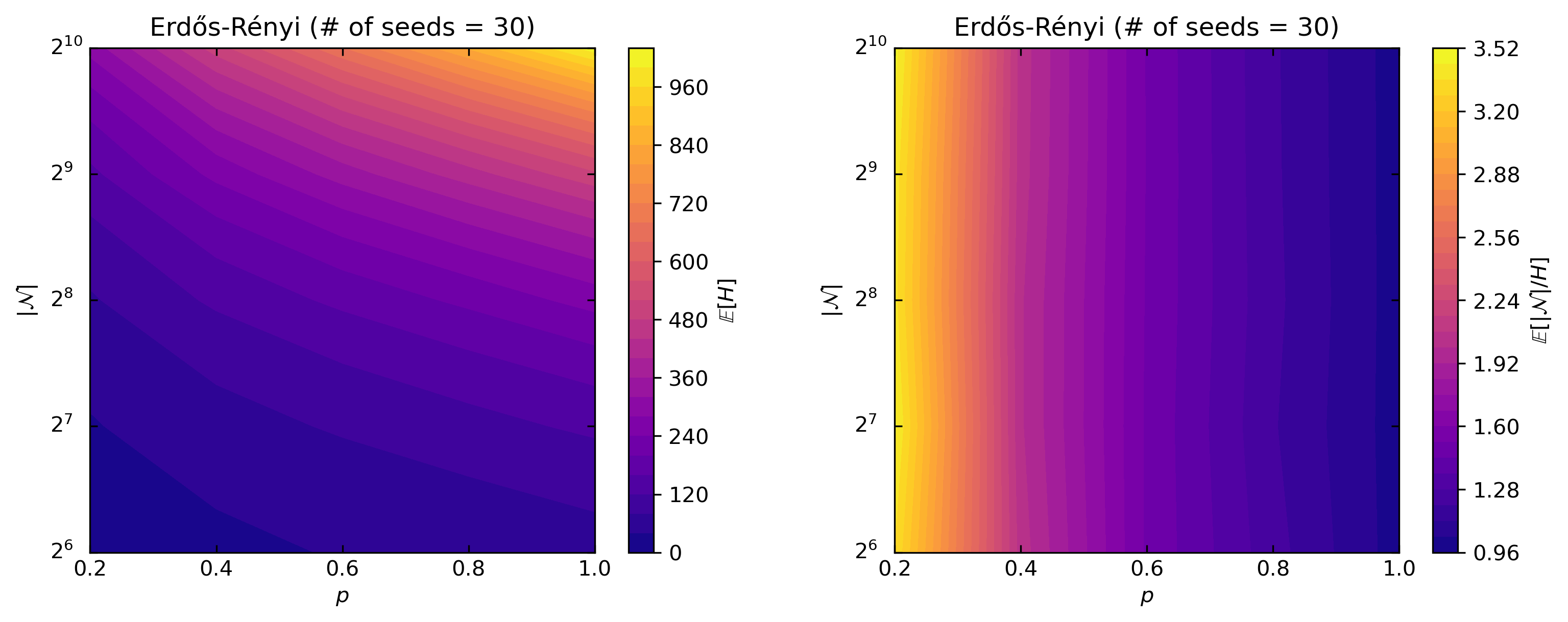}
    \caption{Summary of the defined graph attributes over the investigated ER parameter space. \textbf{Left}: mean height. \textbf{Right}: mean height attenuation.
    Plots span the same size-probability space as Figures \ref{fig:times} and \ref{fig:gains}. Attribute values are color-coded.}
    \label{fig:t-attrs}
\end{figure}

\section{Non-Uniqueness of Minimum Height Layering}

As anticipated in Section \ref{sec:4ward}, there could be multiple minimum height layerings associated to the same DAG. Since the implementation of 4Ward, described in Algorithms \ref{alg:4ward-init} and \ref{alg:4ward-fw}, relies on the specific layering returned by the employed \textit{longest-path} algorithm, it is legitimate to ask what is the impact of different possible partitions on the time complexity of our methodology. In order to address this question, we designed an additional experiment in which we measured the forward pass execution times for different implementations, corresponding to different minimum height layerings, of the same NNs.

Similarly to the experiments described in Section \ref{sec:experiments} and reported in Section \ref{sec:results}, we set $\abs{\mathcal{N}} = 64$, a batch size of 512, and for each density regime $p \in \{0.2, 0.4, 0.6, 0.8, 1.0\}$, we generated 30 ER graphs. Then, for each graph, 10 variants of the same NN were initialized; variants were produced by a modified version of Algorithm \ref{alg:4ward-init} in which the stochastic node reassignment procedure of Algorithm \ref{alg:node-rsg} is executed after the call to the $push\_sources()$ function. The algorithm starts from the layering returned by $longest\_path()$, that is the one in which hidden nodes are placed as close to sinks as possible. Hence, it attempts to randomly reassign hidden nodes to new layers without altering the underlying computational graph. This is done by sampling the new layer of a node from the set of layers that stand between the lowest layer that is higher than those of the node predecessors and the layer of the node itself. Inside the algorithm, $copy()$ clones the set that is passed as an argument while $l_u$ denotes the layer index of node $u$. It is important to highlight that the procedure only changes the order according to which node activations are computed; indeed, forward passes computed feeding several models generated from the same DAG with the same input batch produce exactly the same output.

\begin{algorithm}
\caption{Node reassignment.}\label{alg:node-rsg}
\For{$L \in \{L_2, \dots, L_{H - 2}\}$}{
    $L' \gets copy(L)$\;
    \For{$u \in L'$}{
        preds $\gets \{v : (v, u) \in \mathcal{E}\}$\;
        lower $\gets \max\{l_v : v \in \text{preds}\} + 1$\;
        upper $\gets l_u$\;
        $l \sim \{\text{lower}, \dots, \text{upper}\}$\;
        $L \gets L \setminus \{u\}$\;
        $L_l \gets L_l \cup \{u\}$\;
        $l_u \gets l$\;
    }
}
\end{algorithm}

\begin{figure}
    \centering
    \includegraphics[width=.9\textwidth]{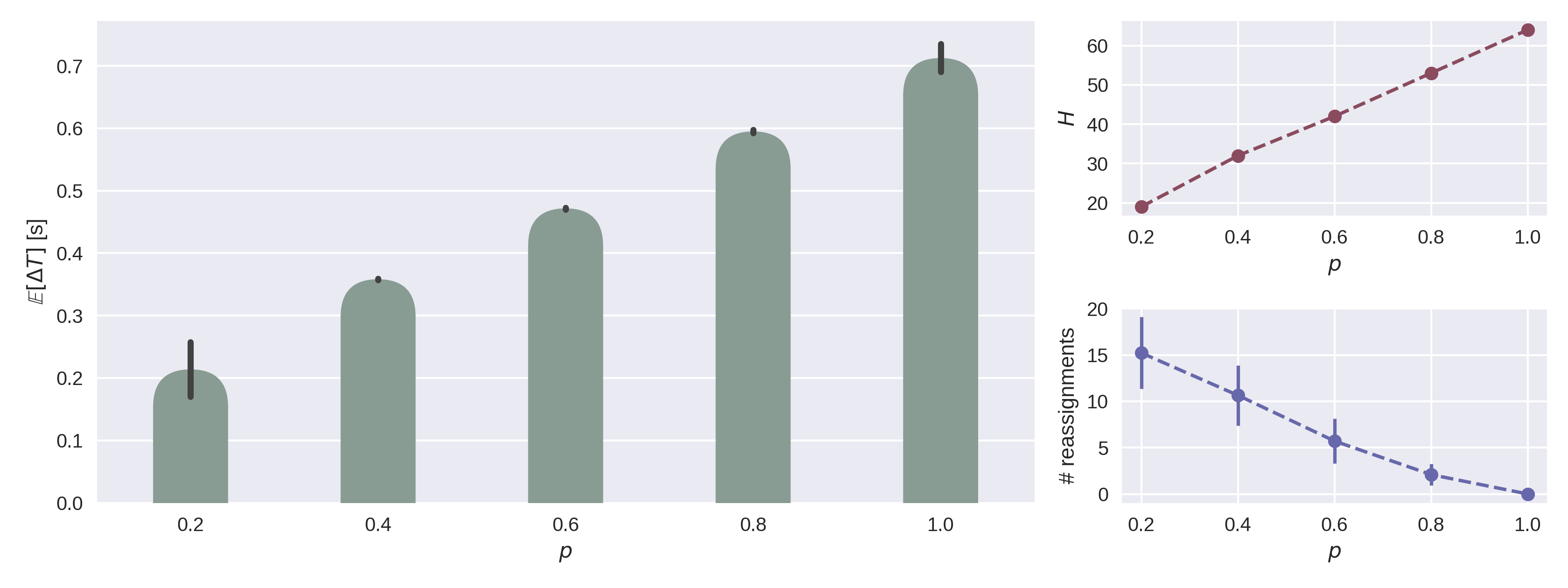}
    \caption{Results of the node reassignment experiment. \textbf{Left}: mean forward pass execution times and standard deviations (black error-bars). Each green bar refers to a specific ER probability/density, $p$. \textbf{Top-right}: height $H$ of the tested computational DAGs. \textbf{Bottom-right}: averge number of node reassignments, with confidence interval (standard deviation), performed by Algorithm \ref{alg:node-rsg} per density regime.}
    \label{fig:rsg-bars}
\end{figure}

The results obtained are shown in the left panel of Figure \ref{fig:rsg-bars}, where, for each density, we report mean ($\mathbb{E}\big[\Delta T\big]$) and standard deviation of the produced distributions of forward pass times (100 batches per measured time). Each distribution has been computed from graphs characterized by the same number of nodes $\abs{\mathcal{N}}$, edges $\abs{\mathcal{E}}$ and layers $H$. This can be accomplished by generating multiple graphs from the same pair $(\abs{\mathcal{N}}, p)$ and keeping only those whose height equals the mean one (pre-computed) that emerges from the generation hyperparameters considered. The standard deviation reported for $p = 1.0$ constitutes our baseline since it is only due to the noise introduced by the hardware and software of the machine on which the experiments have been run; being fully-connected, all the $30 \times 10$ NNs generated in this regime present indeed the same computational DAG and $\mathcal{L}$. When $0.4 \le p \le 0.8$, standard deviations become, counterintuitively, even smaller. The result suggests that the observed variability is again dominated by machine-related noise. We motivate the higher standard deviation of the baseline conjecturing that each matrix multiplication performed could contribute with its own additive noise component to the measured times. And, as shown by the trend depicted in the top-right panel of Figure \ref{fig:rsg-bars}, the number of matrix multiplications performed by a model (i.e., $H - 1$) increases with $p$. The high sparsity regime ($p = 0.2$), instead, presents the highest standard deviation; on average, about $1/4$ of the hidden nodes are in this case reassigned to a new layer (Figure \ref{fig:rsg-bars}, bottom-right panel) by Algorithm \ref{alg:node-rsg}. Overall, the standard deviations are small compared to the values of the computed mean execution times, and, as expected, times, on average, increase with $p$.

\begin{figure}
    \centering
    \includegraphics[width=.49\textwidth]{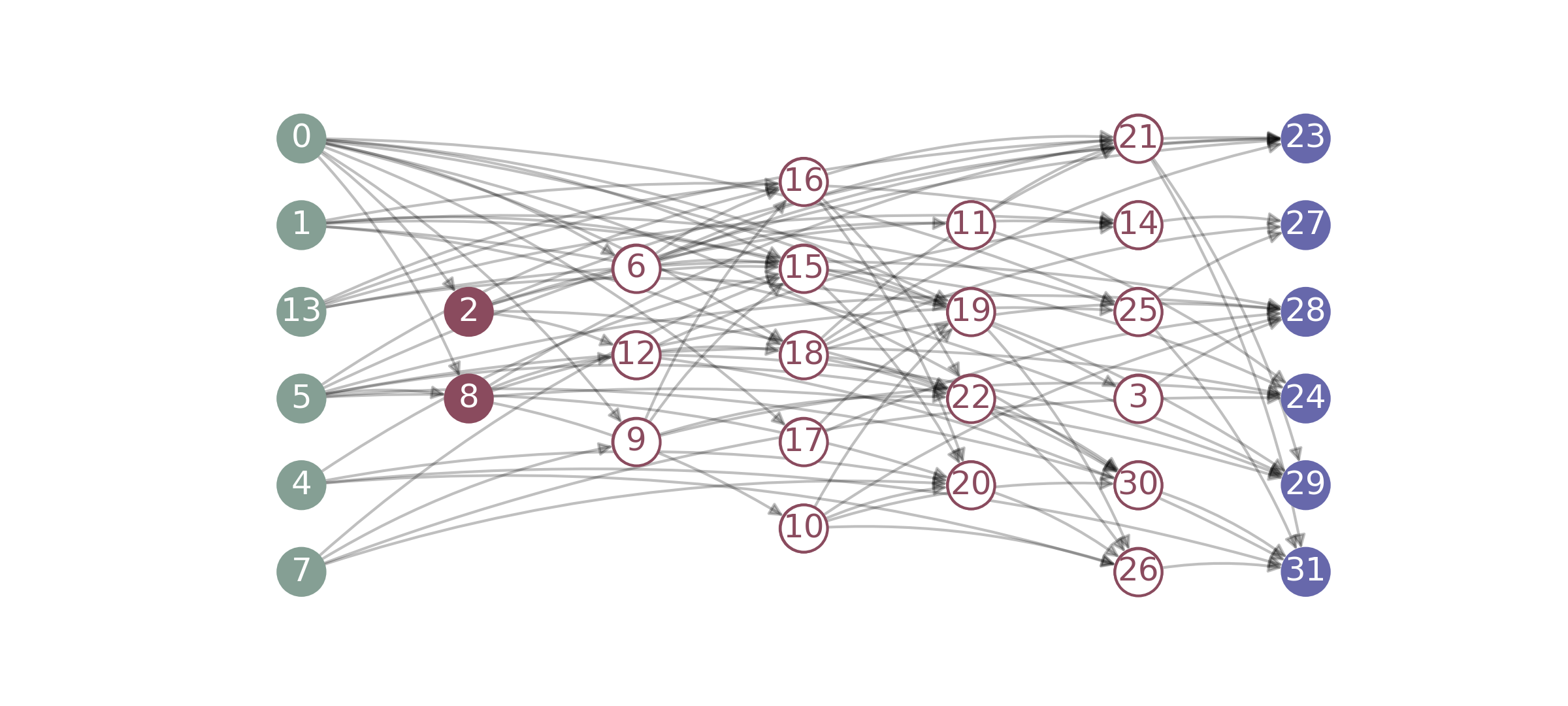}
    \includegraphics[width=.49\textwidth]{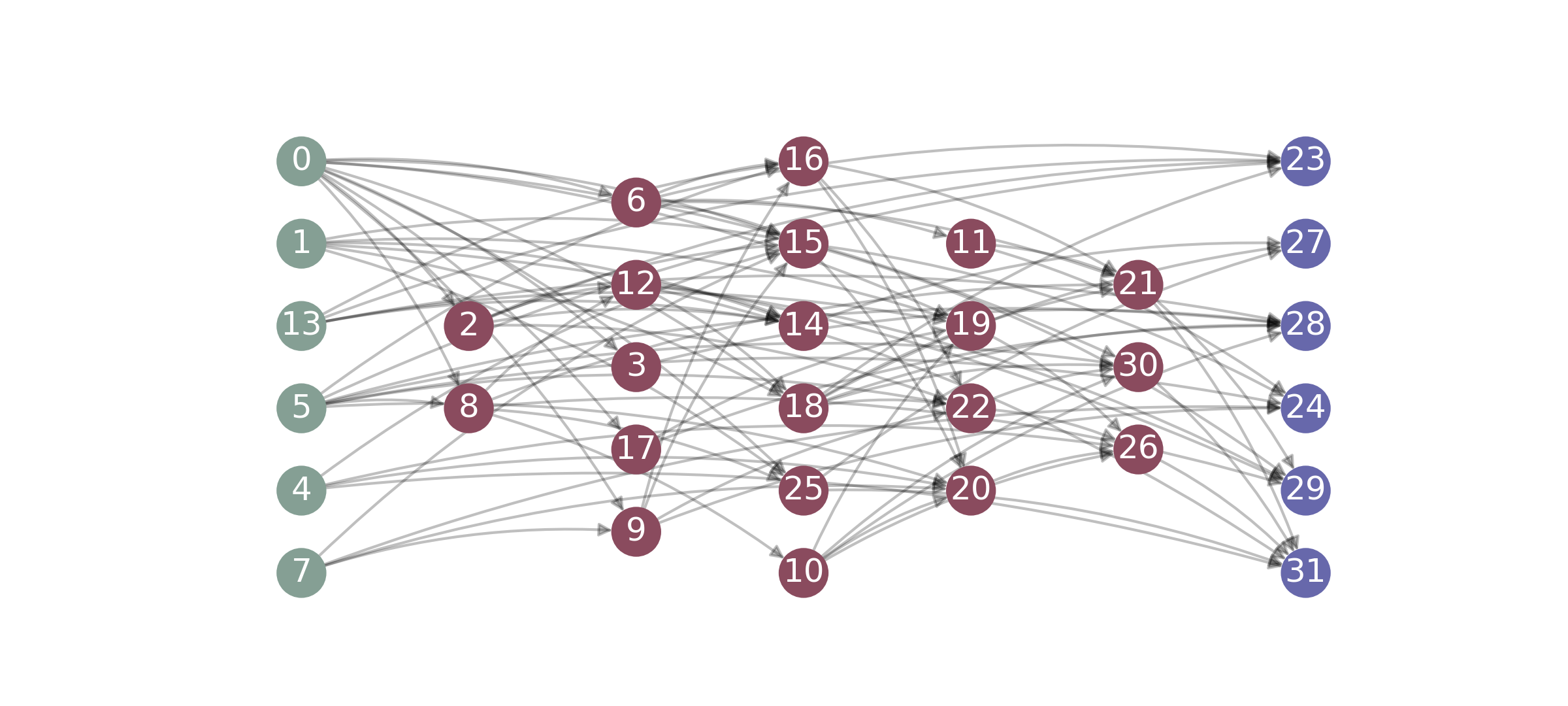}
    \includegraphics[width=.49\textwidth]{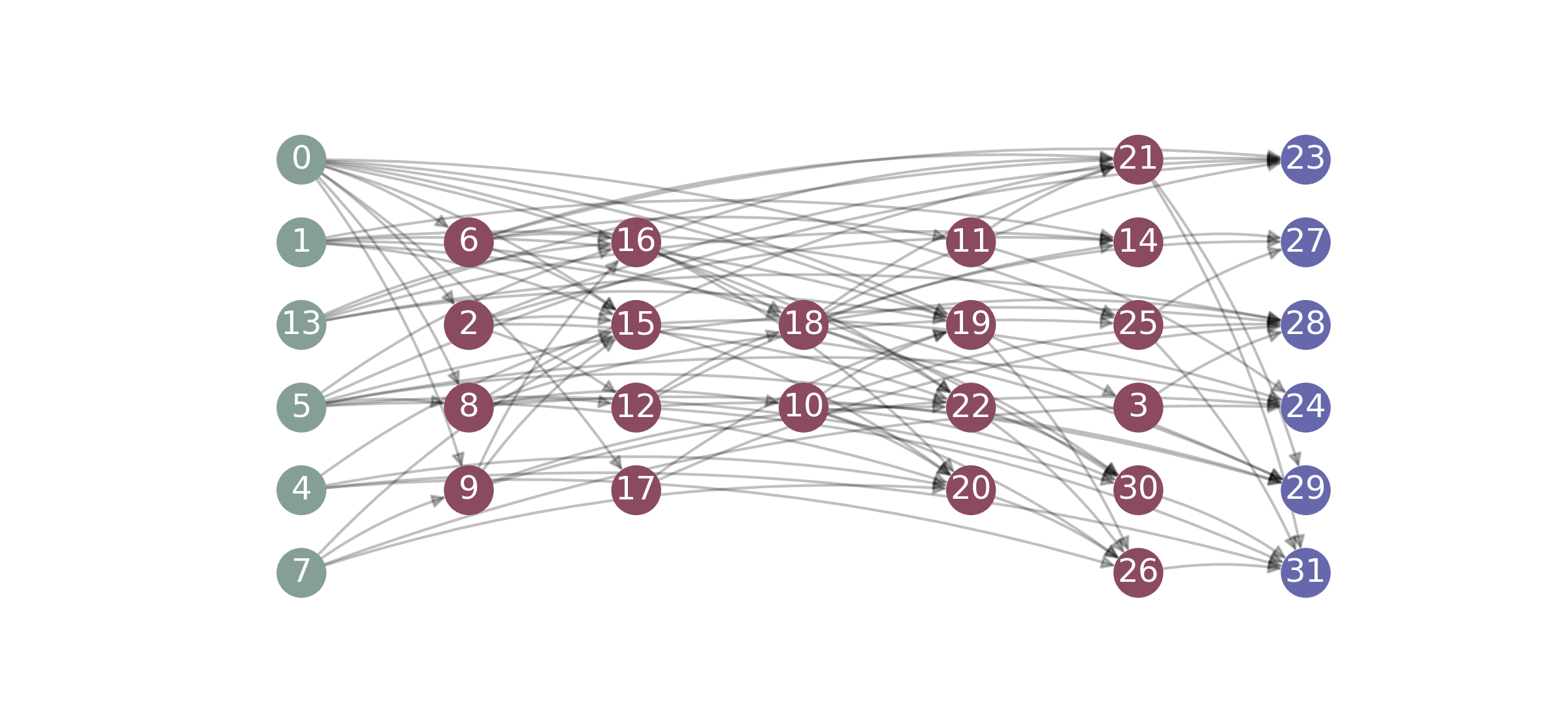}
    \includegraphics[width=.49\textwidth]{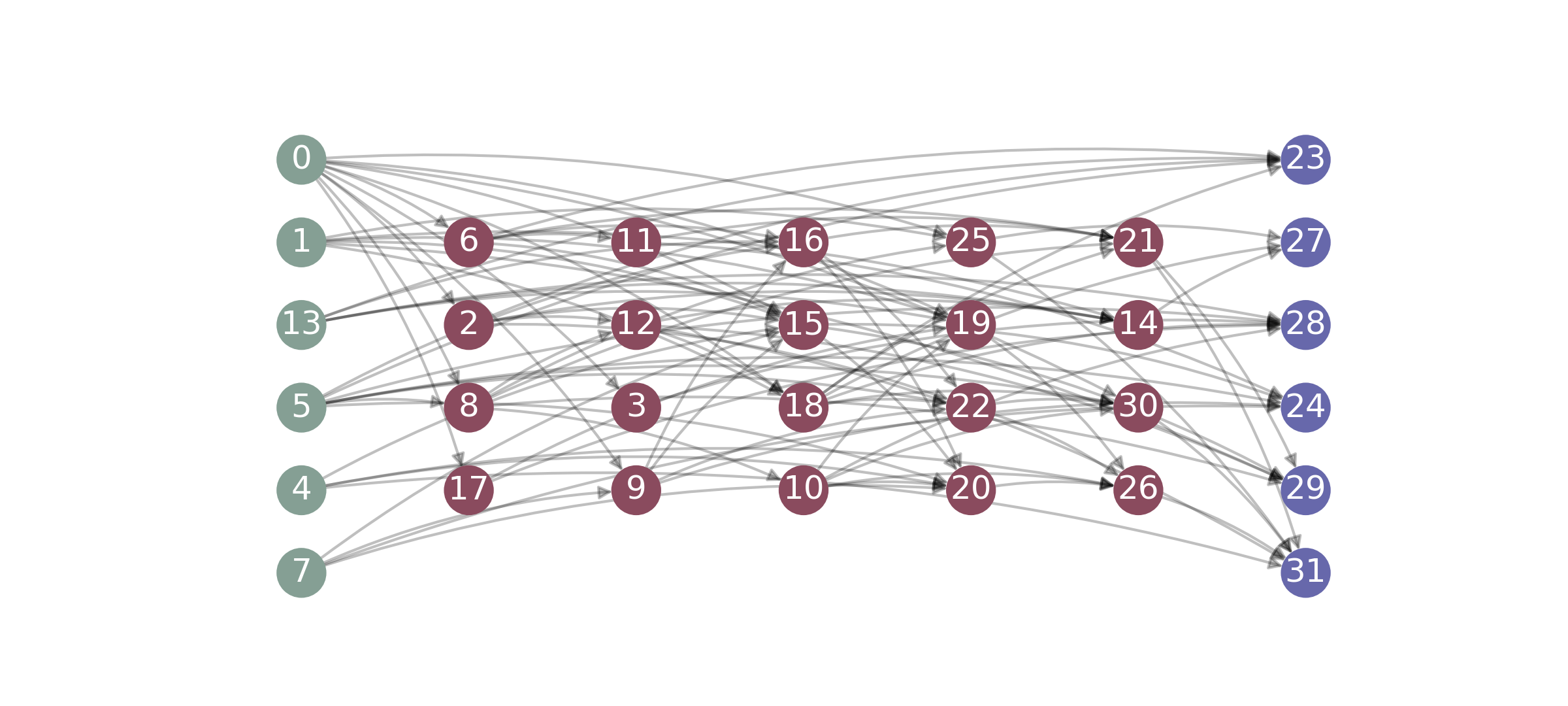}
    \caption{Layering visualizations. All partitions refer to the same 32-node DAG. The top-left layering has been produced through the \textit{longest-path algorithm}; all others were processed by Algorithm \ref{alg:node-rsg}. Sources, hidden nodes and sinks are represented in green, red and violet, respectively. The white vertices in the top-left panel denote nodes that can be moved according to the presented node reassignment procedure.}
    \label{fig:rsg-examples}
\end{figure}

Finally, we specify that not all layerings of minimum height, computable from the same DAG, have the same probability of being generated by Algorithm \ref{alg:node-rsg}; however, we believe that this does not impact what was concluded thanks to the experiments. We conclude reporting in Figure \ref{fig:rsg-examples} a visual example of layerings produced by Algorithm \ref{alg:node-rsg} from the same 32-node DAG/layering pair.

\section{Longest-Path Algorithm}

To ensure comprehensiveness, we report in Algorithm \ref{alg:longest-path} the \textit{longest-path layering algorithm} for partitioning nodes into layerings of minimum height. This algorithm utilizes two sets of vertices, namely $U$ and $Z$, which are initially empty. The variable $l$ represents the label of the layer currently under construction, $\tilde{L}_l$. Whenever a node is assigned to a layer, it is also included in the set $U$, which contains all the vertices already assigned to a layer. On the other hand, $Z$ consists of the vertices assigned to a layer below the current one. To assign a new node $v$ to the current layer, we select it from the vertices that have not yet been assigned to any layer and whose immediate successors are already assigned to the layers below the current one. The final layering $\mathcal{L} = \{L_0, \dots, L_{H - 1}\}$ is obtained by reversing layer indices: $L_i = \tilde{L}_{H - i}$, $0 \le i \le H -1$.

The \textit{longest-path algorithm} possesses two notable advantages: simplicity and linear time complexity. By employing Algorithm \ref{alg:longest-path}, nodes are positioned as close as possible to the sinks.

\begin{algorithm}[H]
\caption{The \textit{longest-path layering algorithm}.}\label{alg:longest-path}
$U \gets \emptyset$\;
$Z \gets \emptyset$\;
$l \gets 1$\;
\While{$U \neq \mathcal{N}$}{
    set $\gets \{v \in \mathcal{N} \setminus U : \{u : (v, u) \in \mathcal{E}\} \subseteq Z\}$\;
    \If{$\textup{set} \neq \emptyset$}{
        $v \sim$ set\;
        $\tilde{L}_l \gets \tilde{L}_l \cup \{v\}$\;
        $U \gets U \cup \{v\}$\;
    }
    \If{$\textup{set} = \emptyset$}{
        $l \gets l + 1$\;
        $Z \gets Z \cup U$\;
    }
}
$\mathcal{L} \gets \{\tilde{L}_H, \dots, \tilde{L}_1\}$\;
\end{algorithm}

 \bibliographystyle{elsarticle-num} 
 \bibliography{cas-refs}





\end{document}